\documentclass[letterpaper, 10 pt, conference]{ieeeconf}  % Comment this line out if you need a4paper

\IEEEoverridecommandlockouts                              % This command is only needed if 
\usepackage{graphics} % for pdf, bitmapped graphics files
\usepackage{graphicx}
\usepackage{amsmath, amssymb}
\usepackage{mathptmx} % assumes new font selection scheme installed
\usepackage{times} % assumes new font selection scheme installed
\usepackage[noend]{algorithmic}
\usepackage[]{url}
\usepackage{subcaption}
\usepackage{textcomp}
\usepackage{siunitx}
\usepackage{tikz}
\usepackage[linesnumbered,ruled]{algorithm2e}
\usetikzlibrary{shapes,arrows}
\usepackage{multirow}
\usepackage{romannum}
\makeatletter
\newcommand{\mypm}{\mathbin{\mathpalette\@mypm\relax}}
\newcommand{\@mypm}[2]{\ooalign{%
  \raisebox{.1\height}{$#1+$}\cr
  \smash{\raisebox{-.6\height}{$#1-$}}\cr}}
\makeatother

\def\inv{\vspace*{-0.15cm}}

\title{\LARGE \bf Analysis of Cellular Feature Differences of Astrocytomas with Distinct Mutational Profiles Using Digitized Histopathology Images}

\author{Mousumi Roy, Fusheng Wang, George Teodoro, Jose Velazqeuz Vega, Daniel Brat and Jun Kong
\thanks{Mousumi Roy, and Fusheng Wang are with the Stony Brook University, Dept. of Computer Science, Stony Brook, NY 11794 ({\tt\footnotesize \{mousumi.roy, fusheng.wang\}@stonybrook.edu}); George Teodoro is with the University of Bras\'{\i}lia, Dept. of Computer Science,  Bras\'{\i}lia, DF, Brazil ({\tt\footnotesize glmteodoro@gmail.com}); Daniel Brat is with the Northwestern University, Dept. of Pathology, Chicago, IL 60611 ({\tt\footnotesize daniel.brat@northwestern.edu}); Jose Velazqeuz Vega and Jun Kong are with the Emory University, Dept. of Biomedical Informatics,  Atlanta, GA 30322 ({\tt\footnotesize \{jose.enrique.velazquez.vega, jun.kong\}@emory.edu}); Funded by NIH K25CA181503, NSF ACI 1443054 and IIS 1350885, and CNPq; The studies involving human subjects were approved by the Emory University IRB.}
 }

\begin{document}

\maketitle
\thispagestyle{empty}
\pagestyle{empty}

%%%%%%%%%%%%%%%%%%%%%%%%%%%%%%%%%%%%%%%%%%%%%%%%%%%%%%%%%%%%%%%%%%%%%%%%%%%%%%%%
\begin{abstract}
Cellular phenotypic features derived from histopathology images are the basis of pathologic diagnosis and are thought to be related to underlying molecular profiles. Due to overwhelming cell numbers and population heterogeneity, it remains challenging to quantitatively compute and compare features of cells with distinct molecular signatures. In this study, we propose a self-reliant and efficient analysis framework that supports quantitative analysis of cellular phenotypic difference across distinct molecular groups. To demonstrate efficacy, we quantitatively analyze astrocytomas that are molecularly characterized as either Isocitrate Dehydrogenase (IDH) mutant (MUT) or wildtype (WT) using imaging data from The Cancer Genome Atlas database. Representative cell instances that are phenotypically different between these two groups are retrieved after segmentation, feature computation, data pruning, dimensionality reduction, and unsupervised clustering. Our analysis is generic and can be applied to a wide set of cell-based biomedical research.
\end{abstract}

%%%%%%%%%%%%%%%%%%%%%%%%%%%%%%%%%%%%%%%%%%%%%%%%%%%%%%%%%%%%%%%%%%%%%%%%%%%%%%%%

\section{INTRODUCTION}~\label{sec:intro}
Large-scale microscopic pathology images have been used in the practice of diagnostic pathology and  understanding of disease mechanisms. Cells are fundamental pathology objects, as they capture rich information on disease characteristics that have been used for diagnostic purposes for over a century. Cell size and shape are controlled by many factors including the genomic,  biochemical, and metabolic status,  signaling networks engaged, physical properties of the plasma membrane, and underlying cytoskeletal properties~\cite{Rangamani}. Therefore, quantitative morphometric analysis may provide information on the genetic or molecular properties of cell populations. Although the prognostic significance of subjective cell features are well known in some diseases, it is often challenging to manually segment and quantitatively analyze cell population features due to their overwhelmingly large numbers in histologic sections. As a result, numerous automated image analysis methods for cell analytics have been proposed~\cite{veta,Xing,histo}. However, there is a lack of complete, self-reliant, modularized processing functions to facilitate the study of cellular phenotypic differences in distinct disease states. To address this, we present a complete and automated cellular feature analysis framework that includes cell segmentation, feature computation, and feature analytics for distinguishing cell populations with distinct molecular signatures. 

As a driving use case for our study, we quantitatively analyze tumor cells from Grade III astrocytomas that are molecularly characterized as either Isocitrate Dehydrogenase (IDH)-mutant (MUT) or wildtype (WT). IDH mutations in infiltrating gliomas identify biologically distinct disease subsets with substantially younger age at presentation, slower clinical progression and longer overall survival compared to those that are IDH wildtype. IDH mutations represent an early, and likely initiating event that is present in more than 80\% of grades II and II astrocytomas and secondary glioblastomas~\cite{Ohgaki}. These genetically and clinically distinct forms of astrocytoma were only recently recognized and their morphologic differences have not been described, providing an excellent test case for framework development.  

\section{METHOD}~\label{sec:methodology}
Our cell analysis workflow consists of a sequence of steps, including tumor region annotation, tumor region image extraction, image stain normalization, stain color deconvolution, cell segmentation,  cellular feature computation, and  feature comparison analysis with representative cell retrieval.% with feature analysis.

\subsection{Tumor Cell Segmentation} ~\label{ssec:seg}
\begin{figure}[tb!]
\centerline{
\includegraphics[width=\linewidth]{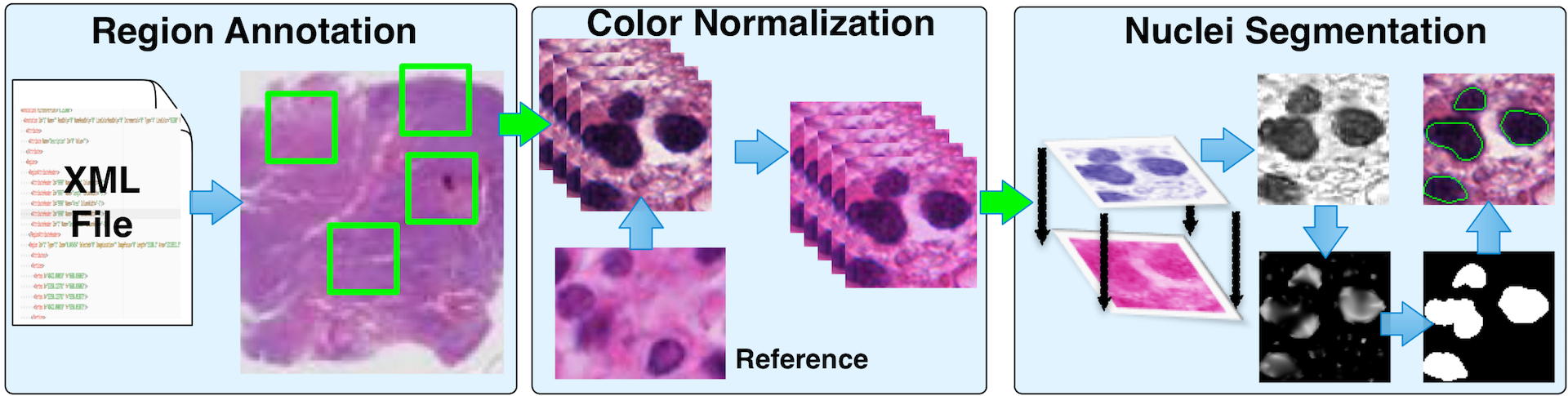}
}\inv
\caption{Overall schema of cell analysis is presented.}~\label{fig:segmentation}\inv\inv\inv\inv\inv
\end{figure}

The overall schema for cell segmentation is  presented in Figure~\ref{fig:segmentation}. As not all tissue regions in a  whole-slide microscopy image are tumor regions of interest, neuropathologists manually select Regions of Interest (ROIs) from each slide~\cite{imagescope}. The resulting annotation results are captured and exported in xml files. Next, our analysis module programmatically reads xml files and uses openSlide API to retrieve the annotated image regions from whole-slide images~\cite{openslide}. %We generate 200 images of tissue regions from 50 patients equally from two molecular groups.

%\subsection{Tumor Cell Segmentation}~\label{ssec:segmentation}
%Next, we segment all tumor nuclei from each image. As it is not unusual to observe large variations in color and illumination of slides due to different slide preparation, staining, and scanning processes, it is necessary to normalize slide color that can benefit the robustness of the following segmentation analysis. 
To begin the cell segmentation analysis, we normalize colors of all images to mitigate the impact of stain variation on the follow-up segmentation. Aiming for color channel correlation minimization, we convert each image from the RGB to LAB color space where each channel is mapped by mean and standard deviation of the corresponding image channel from target image, respectively~\cite{Magee}. The transformed color channels are then mapped back to RGB space~\cite{Reinhard}. All slides are stained by Hematoxylin and Eosin (H\&E) stain, resulting in regions of nuclei and cytoplasms in purple and pink. %This normalization step produces significant intensity contrast between distinct histopathology components, indicating a promising segmentation result. 
To target signals useful for segmentation, we decouple these two distinct stains from the original color image by Lambert-Beer's law~\cite{Ruifrok}. %We deconvolve color components by Lambert-Beer's law~\cite{Ruifrok}. It characterizes the relationship between the intensity of light entering a specimen $I_i$ and that through a specimen $I_o$ as: $I_o = I_i e^{(-A b)}$ where $A$ and $b$ are the amount of stain and the absorption factor, respectively. The resulting Optical Density (OD) is  defined as: $OD = -\log(I_o / I_i) $. We define the un-mixing $3\times 3$ matrix $U$ with its three columns for OD values associated with the red, green, and blue channel for Hematoxylin, Eosin, and a Null place-holding stain. Denoting $C(m,n)$ is a $3\times 1$ vector representing amounts of three stains at pixel $(m,n)$, the OD levels for red, green, and blue channel $L$ would be $L = UC$. As a result, an orthogonal representation of the stains can be written as $C = U^{-1} L$, where $U^{-1}$ is the color deconvolution matrix.
%With the deconvolved Hematoxylin image channel associated with nuclei, we only focus on Hematoxylin image channel for nuclei segmentation in the following process.
As nuclei in cells are highlighted by Hematoxylin, we next process the deconvolved Hematoxylin image channel for further segmentation. 

We explore the resulting Hematoxylin image channel and find that pixels tend to have gradually decreased Hematoxylin stain signal intensity as they approach to nuclear centroid. To enhance signal contrast for more accurate segmentation, we determine the likelihood of a given pixel in cell regions by analyzing eigenvalues of the Hessian matrix from local image neighbors~\cite{Frangi}. Given the prior knowledge about cell shape, we search for circular structures in Hematoxylin channel and compute likelihood for cellular pixels based on geometric structures characterized by the neighboring pixel intensity profiles. For any arbitrary pixel at $(x_0, y_0)$, its local image intensity change can be represented by Taylor expansion:
\begin{align}
&h_{\sigma}(x_0+\delta x, y_0+\delta y) = h_{\sigma}(x_0, y_0) + (\delta x, \delta y) D\left(h_{\sigma}\right)|_{(x_0, y_0)}\\ \nonumber
& + (\delta x, \delta y) D^2\left(h_{\sigma}\right)|_{(x_0, y_0)} (\delta x, \delta y)^T +  \mathcal{O}((\delta^3) \nonumber
\end{align}
\noindent where $h_{\sigma}$ is the convolution of  the deconvolved Hematoxylin image channel $h$ and a Gaussian filter $G_\sigma$ with standard deviation $\sigma$; $D^2\left(h\right)$ is Hessian of $h$ that is symmetric and thus diagonalizable with two resulting eigenvalues $\lambda_i, i=1,2$. Due to the  Hematoxylin channel property, it is straightforward to have $0\ll \lambda_1 \lessapprox \lambda_2$ for pixels within cells. To improve the intensity contrast between cells and background, we use the following cell enhancement function~\cite{Frangi}: $f(h_\sigma(x,y), \alpha, \beta) = \left(1-exp\left(-\frac{(\lambda_1/\lambda_2)^2}{2\alpha^2}\right)\right)\left(1-exp\left(-\frac{\lambda_1^2+\lambda_2^2}{2\beta^2}\right)\right)$ where $\alpha$ and $\beta$ are sensitivity parameters for two product terms. As cell size is variable, we take the maximum response with the optimal scale $\sigma^\ast$. The resulting enhanced likelihood map is further processed with hysteresis thresholding~\cite{Hancock}. In the post-processing step, we exclude the resulting candidate either too small or too bright to be a cell. For those candidates with internal holes, we fill up holes. The cellular boundary is smoothed by a low pass filter in the end. 

\subsection{Feature Computation}~\label{ssec:feature}
To remove all erroneous results generated by machine analysis, the resulting cellular boundaries are manually corrected by domain experts before they are used for feature computation. Manual corrections included annotations for mitotic figures, apoptotic nuclei, corpora amylacea, neurons, and endothelial cells.  Abnormally segmented cells, such as those clumped cells and partial cells on image edges, are removed. After this validation and correction process by human experts, we proceed with cellular feature analysis by computing % nine% 
quantitative features related to cell size, shape, intensity, and hyper-chromaticity. Specifically, for each segmented cell, \textit{Area, Perimeter, Max Distance}, and \textit{Equivalent Diameter} are computed to represent cell size;  \textit{Eccentricity, Circularity}, and \textit{Extent} are derived to describe cell shape; \textit{Intensity Standard Deviation}, and \textit{Intensity Entropy} are used to characterize Hyper-chromaticity; \textit{Intensity Mean} is used to represent cell pixel intensity.
%In aggregation, 50,588 nuclei are automatically analyzed for feature computation, with 26,871 and  23,717 from MUT and WT, respectively.
%segmented and computed with nuclear features, 26,871 from IDH-mutant AAs and the remaining 23,717 from IDH-wt AAs.

\subsection{Feature Analysis}~\label{ssec:representative}
To visually perceive cellular feature difference, we project data to a lower dimensional subspace with Multi-Dimensional Scaling (MDS)~\cite{modern} after feature normalization. %MDS is a non-linear dimensionality reduction method that preserves data similarity distances. 

\begin{figure}[tb!]
\centerline{
\includegraphics[width=\linewidth]{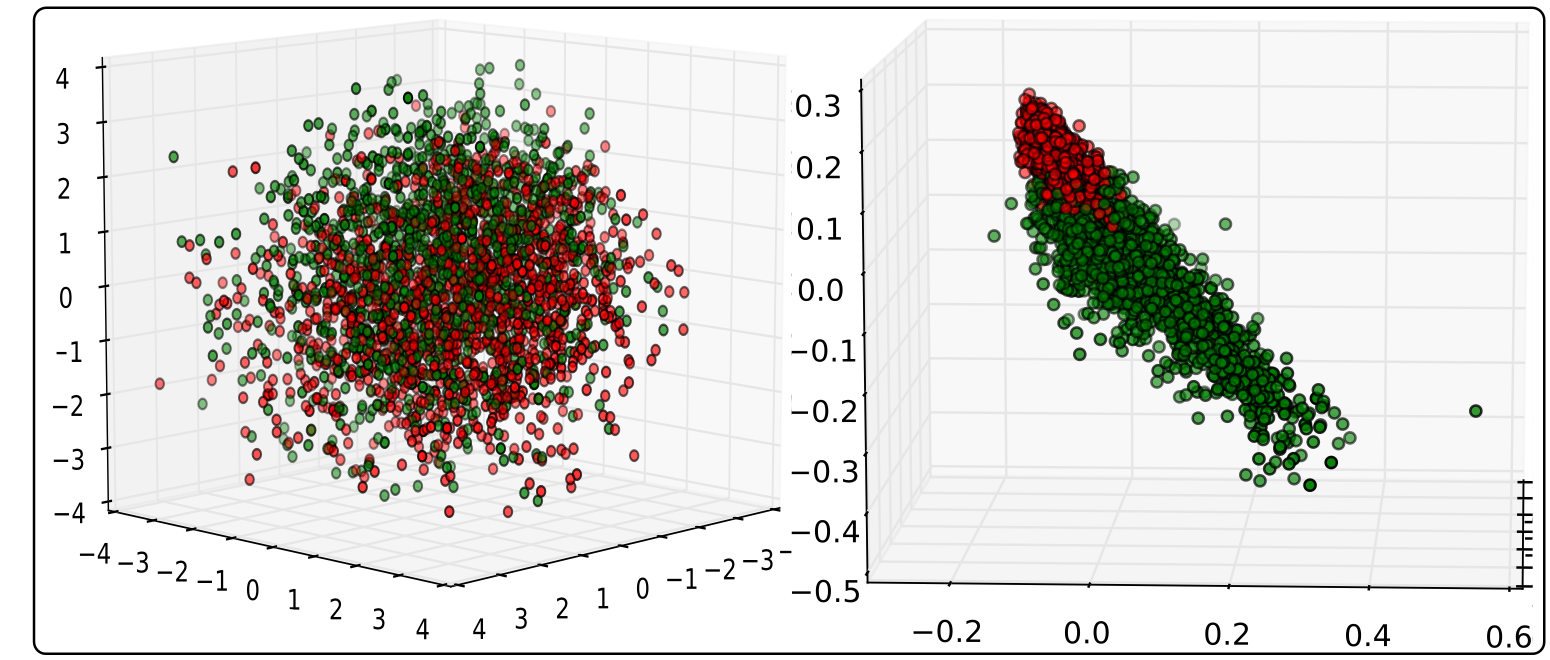}
} \inv %\inv \inv
\caption{MDS derived scatter plots (Left) before and (Right) after data pruning are demonstrated.}\label{fig:MDS} \inv\inv\inv
\end{figure}

MDS is a non-linear dimensionality reduction method that detects the underlying dimensions of data by finding the similarity or dissimilarity between pairs of data. 
%detects the underlying dimensions by finding the similarity or dissimilarity of the objects. 
%To apply MDS on the dataset 'I', 
%Distance function $d_{ij}$ is defined and stored in the dissimilarity matrix $\Delta = \begin{bmatrix}\delta_{i,j}\end{bmatrix} \text{ for all } i,j \in 1,...,I $. 
Given the dissimilarity matrix $\Delta= \begin{bmatrix}||y_i - y_j|| = \delta_{i,j}\end{bmatrix}$ %\text{ for all } i,j \in 1,...,I $, 
derived from the original feature space with dimensionality of $p$,  our goal is to find $\{x_i$% ,..., $x_I$ 
$\in \mathbb{R}^d\}$ with $d < p$ such that $||x_i - x_j|| \approx \delta_{i,j}$. $x_i\in \mathbb{R}^d$ is the low dimensional data representation of $y_i\in \mathbb{R}^p$ in a higher dimensional space. %\text{ for all } i,j \in 1,...,I $. %where $|| . ||$ is a vector norm which is Euclidean distance for Classical MDS. 
MDS generalizes the optimization method by minimizing the cost function: % called ``Stress'' (S), which is a residual sum of squares: 
$S = \sum\limits_{i \neq j}(\delta_{i,j} - ||x_i - x_j ||)^2)$. %^{1/2}$. %After dimensionality reduction with MDS, we next apply Linear discriminant analysis (LDA) to two populations of projected data that are further projected to one dimension perpendicular to the final two-class decision boundary. 
For visual assessment, we apply MDS and project data to a lower space (i.e. $d=3$) with MDS and present data in Figure~\ref{fig:MDS} (Left) where data from MUT and WT are observed to be severely overlapped. As there is a large number of cells from multiple cell populations with large phenotypic and biological variation in each image, severe noise is embedded in data representation as suggested by the nonlinear dimensionality reduction method MDS. Therefore, it is more promising to find true cell phenotypic feature signatures by data pruning than exhaustively formulating additional cell feature representations in this scenario. Following this idea, we measure discriminating power of cells in an ensemble classification manner and robustly find representative instances distinctive across two groups. %Only those representative instances from each class are retained.

\subsubsection{Data Pruning}~\label{sssec:subsubhead1}
We prune data with a mixed collection of generative and discriminative classifiers at this stage: Logistic Regression (LR), Random Forest (RF), AdaBoost (AB), Naive Bayes (NB), Quadratic Discriminant Analysis (QDA) and Neural Net (NN)~\cite{Log,ada}. %~\cite{Log,random,ada}. 
%Of these methods, they can be divided into two categories, generative, and discriminative method. 
Generative models are used to model the distribution of individual class. These methods derive the posterior probability on class label $y$ given observation $x$ using Bayes rule after modeling $p(y)$ and $p(x|y)$: 
$\underset{y}{\operatorname{arg\:max}}  p(y|x) = \underset{x}{\operatorname{arg\:max}}  p(x|y)p(y)$. %NB, QDA are examples of such generative models. 
By contrast, discriminative methods predict the class label $y$ from the training example $x$, by evaluating: $f(x) = \underset{y}{\operatorname{arg\:max}} \: p(y|x)$. %These models provide classification splits in order to separate classes. 
%LR, RF are examples of discriminative models.

%LR is a predictive analysis which uses logistic function to define the hypothesis $h_\theta(x)$ as follows~\cite{Log}: $h_\theta(x) = g(\theta^T x) = \frac{1}{1 + e^{- \theta^T x}}$. RF is an ensemble learning method which constructs a multitude of decision trees at training time and then outputs the class which is the mode of the classes~\cite{random}. AB is used with other learning algorithms to improve their performance~\cite{ada}. 

We use ensemble classifiers to vote for discriminating cell instances in a robust way~\cite{Rokach}. We have trained all these classifiers and classified data into two groups: MUT and WT. Five-fold  cross validation is used to mitigate over fitting problem with the training data. We prune data iteratively in a few successive steps. First, those instances that are correctly classified by any of the above classifiers are kept. In the next iteration, we train classifiers with the remaining data and only keep those instances correctly classified by at least two updated classifiers. We increment the number of classifiers that produce correct classification results in each step. Following this method, the remaining dataset correctly classified by at least five classifiers are retained in the last iteration. We stop from increasing the number of correct classifiers further as we notice the classification accuracies of all classifiers get saturated up to this iteration.  Finally, those confusing instances that are misclassified by any of the classifiers are removed.  
%Although, the current dataset is labeled for each group, the trained classifiers provide the distinction criteria between two groups. We can use the trained classifiers for future unlabeled data.
%instead of any single classifier to find the cell instances with the least discriminating information, as the classification result from individual classifier varies.

\begin{algorithm}[tb!]
\scriptsize
\begin{flushleft}
\caption{Algorithm for representative cell retrieval}
\label{alg}
{\bf Input:} $D$: \emph{data feature matrix}; N: number of rows in $\it{D}$; $F \leftarrow 9; C$: cluster number \\
{\bf Output:} $P_i$: \emph{cell panel for cluster $i$ of each group}
\begin{algorithmic}[1]     
    \FORALL {$\mathrm{groups} \quad G \in (0,1) $} 
    \STATE $D_0 \leftarrow D[G] ; \mathrm{count} \leftarrow 0$
	 \FORALL {$c \in (0,1,..,C)$} 
     \FORALL {$i \in (0,1, ..., N-1) $} 
         \STATE  $\mathrm{count}[c] \leftarrow \mathrm{count}[c] + 1 $ 
        \ENDFOR
	\STATE $\overline{c} \leftarrow avg(D_0, \mathrm{count}) \qquad //\overline{c}$:  cluster centroid 
   	\FORALL{$i \in (0,1,...,N-1)$}
    	\STATE $\delta[i] \leftarrow \mathrm{euclidDist}(\overline{c},D_0[i])$
	\ENDFOR 
     \STATE append($D_0, \delta$)
     \STATE $D_0^s \leftarrow sort(D_0, \delta)$
     \STATE  // Find the $w_{max}$ and $h_{max}$ of MBR from 100 cells 
     \STATE $w_{\mathrm{max}} \leftarrow 0;  h_{\mathrm{max}} \leftarrow 0$
     \FORALL{$j \in (1,2,...,100)$}
     	\STATE $ x_{{\min}} \leftarrow \min(N_B[0]) ; y_{{\min}} \leftarrow \min(N_B[1]) $
        \STATE $ x_{{\max}} \leftarrow \max(N_B[0]) ; y_{{\max}} \leftarrow \max(N_B[1])$
        \STATE $w \leftarrow x_{{\max}} - x_{{\min}} ; h \leftarrow y_{{\max}} - y_{{\min}}$
        \STATE $w_{{\max}} \leftarrow \max(w,w_{{\max}}) ; h_{{\max}} \leftarrow \max(h,h_{{\max}}) $  
 	\ENDFOR
     \STATE $w_{\text{final}} \leftarrow w_{{\max}} + x_{\text{margin}} ; h_{\text{final}} \leftarrow h_{{\max}} + h_{\text{margin}}$ 
     \FORALL{$j \in (1,2,...,100)$}
%     	\STATE $ x_{\text{min}} \leftarrow min(nu_{\text{boundary}}[0]) ; y_{min} \leftarrow min(nu_{\text{boundary}}[1])$
     	\STATE $ x_{{\min}} \leftarrow \min(N_B[0]) ; y_{\min} \leftarrow \min(N_B[1])$
        \STATE $ N_{\text{MBR}}[j] \leftarrow crop(I, [x_{{\min}}, y_{{\min}}, w_{\text{final}}, h_{\text{final}}])$
      \ENDFOR  
	\FORALL { $j \in (1,2,.. ,100)$ }
     		\STATE $P_i$: image panel generated by adding $N_{\text{MBR}}[j]$ 
     \ENDFOR
     \ENDFOR
     \ENDFOR
\end{algorithmic}
\end{flushleft}\inv
\end{algorithm}

\begin{figure}[b!]
\centerline{
\includegraphics[width=\linewidth]{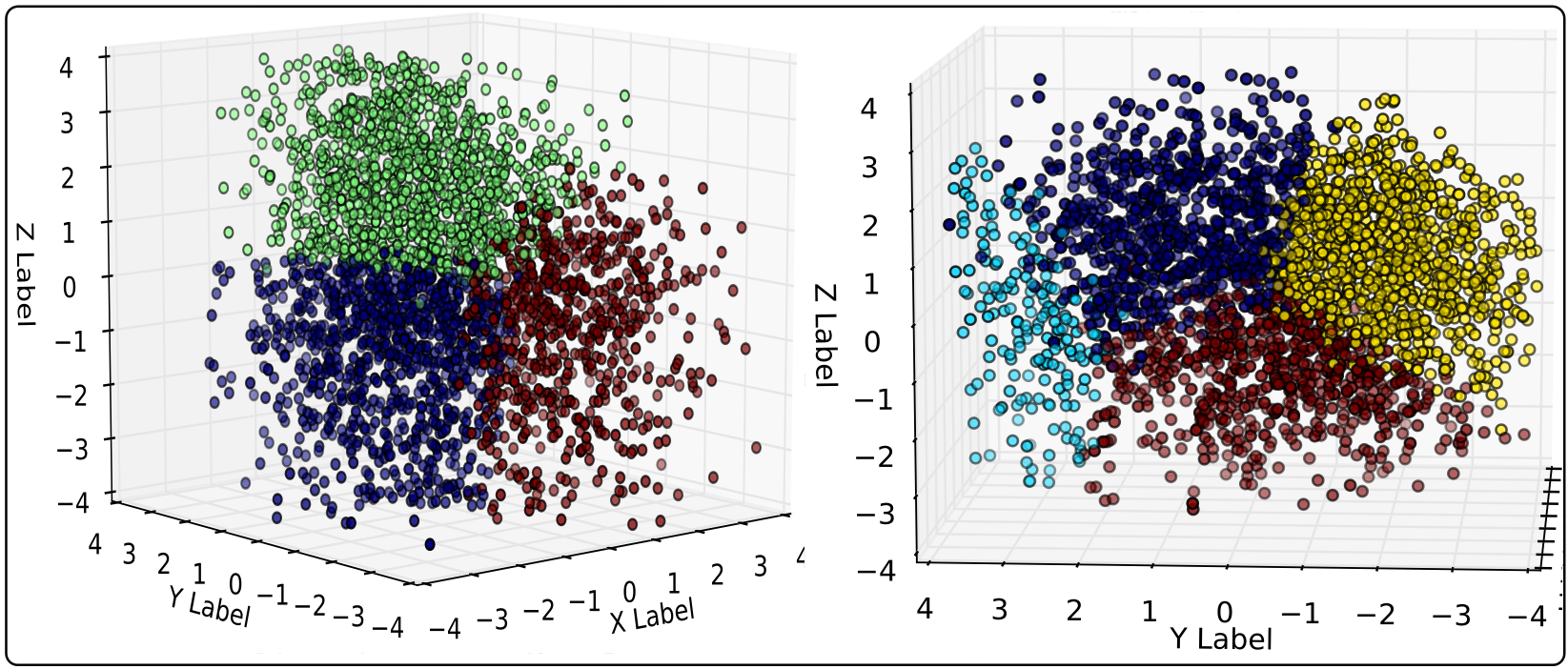}
} \inv %\inv \inv
\caption{Scatter plots of clustered data from (Left) MUT and (Right) WT group are presented.}\label{fig:cluster} \inv \inv \inv
\end{figure}

\subsubsection{Data Clustering Analysis}~\label{sssec:subsubhead2}
With the retained data, we next analyze the hidden structure of group MUT and WT and find the most representative cell instances of each group. Specifically, we recover the structure of the unlabeled data of each group and partition them into $K$ sets by K-means clustering~\cite{Celebi}. %We use K- means method to partition data points %$(x_1, x_2, ... x_n)$ which are 9-D real vectors of each group, 
%into $K$ ?sets %S = {$S_1, S_2,..., S_k$},
%in an unsupervised manner, so as to minimize the objective function defined as the within-cluster sum of squares~\cite{art}. 
We change $K$ from 1 to 10. For each value, we compute the Sum of Squared Errors (SSE). %$J = \sum\limits_{j=1}^{K} \sum\limits_{i=1}^{N} \|x_i^{j} - c_j\|^2$, where $K$ is the number of clusters and $N$ is the number of data points. %{\bf Then, we have plotted a line chart of the SSE for each value of k, which looks like an arm and the "elbow" on the arm is the optimal k.} 
Overall, SSE  decreases as we increase $K$. We aim to choose as a small $K$ as possible that produces a sufficiently low SSE. The ``elbow'' point usually represents the point after which SSE does not reduce much by increasing cluster number. We detect the ``elbow'' point where the graph begins to flatten significantly (i.e. the decrease percentage of SSE is $< 15\%$ in our case)~\cite{elbow}. Following this approach, we find $K=3$ and $K=4$ as the optimal cluster number for MUT and WT group, respectively. Clustering analysis for each group is next performed with the optimal $K$ value.

%\inv \inv \sinv
\subsubsection{Retrieval of Representative Cells}
With data clusters from each group, representative cells are found with Algorithm~\ref{alg} that computes the centroid of each cluster and identifies the first 100 most representative cells based on their Euclidean distances to the cluster centroid. In Algorithm~\ref{alg}, variable $\delta$ represents a distance vector that captures Euclidean distances from data points to the cluster centroid; $I$ is the image for cell analysis; $D_0$ and $N_B$ capture the cellular features and boundary coordinates. Representative cells capturing group difference signatures are extracted from corresponding images $\{I\}$ by fitting minimum bounding rectangles from $\{N_B\}$ and assembled in a $10\times 10$ cellular panel for each group, respectively.

\begin{figure}[tb!]
\centerline{
\includegraphics[width=\linewidth]{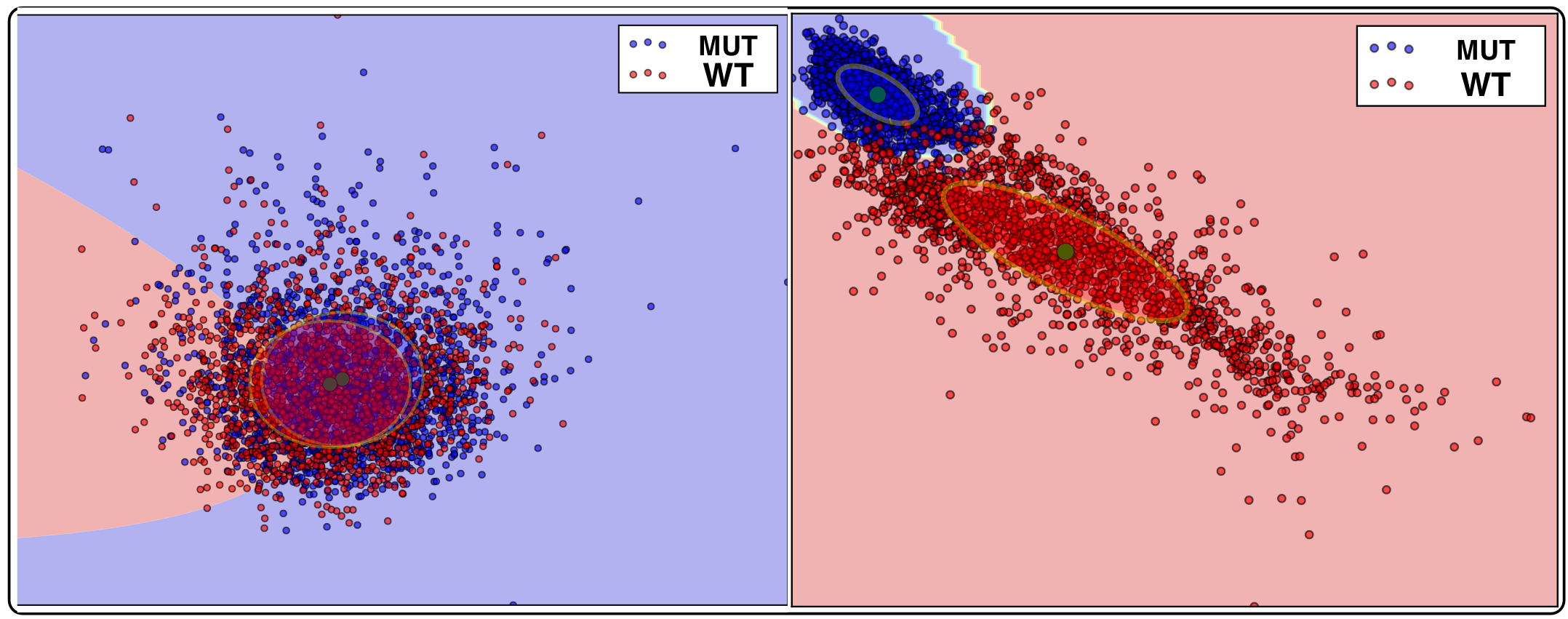}
} \inv %\inv \inv
\caption{Decision boundaries from quadratic discriminant analysis (Left) before and (Right) after data pruning are presented.}\label{fig:qda} \inv \inv\inv
\end{figure}

\section{EXPERIMENTAL RESULTS}~\label{sec:expres}
As a driving use case, we test the developed analysis pipeline with histopathology images of astrocytoma tissues (i.e. a malignant brain tumor) from The Cancer Genome Atlas database. This dataset includes 200 images ($1024\times 1024$) of tumor regions from 50 patients equally from two molecular groups: Isocitrate Dehydrogenase (IDH) mutant (MUT) and wildtype (WT). In aggregation, 50,588 nuclei are automatically analyzed for feature computation, with 26,871 and  23,717 from MUT and WT, respectively. 

To visualize data separability, we project data to a lower three-dimensional feature space with MDS and produce scatter plots with data before and after pruning in Figure~\ref{fig:MDS}. It is clear that the mapped feature data from two groups are highly clumped before data pruning, whereas post-pruning data present substantially improved separability between two groups. In Figure~\ref{fig:qda}, we further visualize the decision boundary with the projected data from the two groups with QDA. The mode of the distribution of each group is specified by a confidence ellipse, with blue and red region for MUT and WT, respectively. %The figures represent those nuclei instances, which are  most different in feature wise across the two groups. 
%The classification accuracy after data pruning on testing data is $100\%$ for LR, AB, NB, QDA, NN, and $93\%$ for RF. 
The dataset after data pruning process includes 20,653 nuclei instances (i.e. about 41\% of the original data) of which 8,257 and 12,396 cells are from group MUT and WT, respectively.

\begin{table}[b!]
\small
\centering
\caption{Mean and standard deviation of most discriminating cellular features for distinguishing cells in group MUT from WT are presented.}~\label{table:mean_std}
\inv\inv
%\smallskip\noindent
\resizebox{\linewidth}{!}{
 \begin{tabular}{|c | c c c c|} 
 \hline
 Group & Area & Perimeter & MeanIntensity & MaxDistance\\ %[0.5ex] 
 \hline%\hline
 MUT & 448.28$\pm$162.88  & 76.36$\pm$14.96 & 52.50$\pm$16.00 & 28.33$\pm$6.16 \\ 
 \hline
 WT & 198.97$\pm$82.12 & 54.09$\pm$13.61 & 71.63$\pm$24.29 & 22.09$\pm$6.43 \\%[1ex] 
 \hline
\end{tabular}
}
\end{table}

\begin{figure}[tb!]
\centerline{
\includegraphics[width=\linewidth]{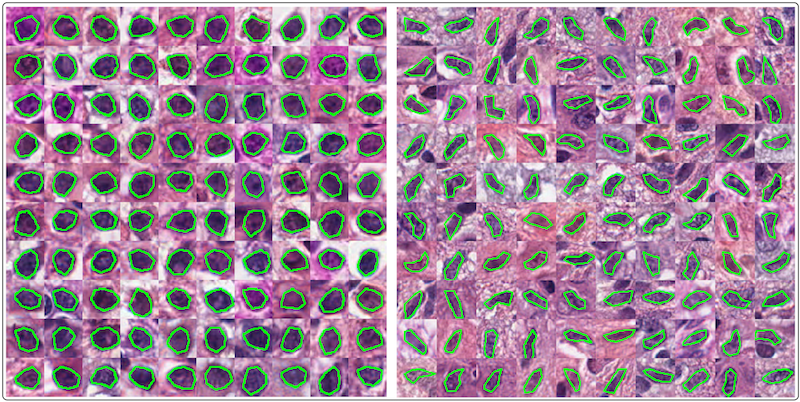}
}\inv %\inv %\inv
\caption{Two representative panels of retrieved cells from (Left) MUT and (Right) WT group are presented.}~\label{fig:cell} \inv\inv\inv\inv\inv\inv
%\caption{Cellular panels from (Top) MUT and (Bottom) WT are presented, respectively.}~\label{fig:cell} \inv \inv \inv\inv\inv\inv\inv
\end{figure}

Table~\ref{table:mean_std} presents mean and standard deviation of the most discriminating features of cells between MUT and WT after data pruning. %From the histogram and other statistical values, like mean($\mu$) and standard deviations($\sigma$) of all features across the two groups, 
After statistical analysis and feature histogram study, we find \textit{Area} and \textit{Perimeter} are most discriminating cellular features between group MUT and WT. %The mean nuclear area for MUT(448.28) is much higher than that of WT (198.96).  
%\begin{figure}[t!b]
%\centerline{
%\includegraphics[width=\linewidth]{Fig/histogram.png}
%}
%\caption{Histogram of Area(left) and Circularity (right)}\label{fig:hist}
%\end{figure}

To visualize representative cells that carry discriminating information from each group, we cluster each group with K-means algorithm with optimal cluster number. We present scatter plot for each group in Figure~\ref{fig:cluster}. It is observed that relatively well compact clusters are formed with clear cluster separation for each group. Additionally, cells from same clusters have very similar feature representations and those from different clusters present distinct feature signature.

After clustering data for each group, we retrieve the first 100 representative cells based on the Euclidean distance to each cluster centroid for each group. A minimum bounding rectangle is found for each cell. The resulting cell image regions are extracted from the associated images and assembled into a 10 by 10 image panel. We repeat this process for each specific cluster of each group. In this way, we can facilitate the visualization of representative cells that capture discriminating information for differentiation of cell populations with distinct genetic or molecular properties. In Figure~\ref{fig:cell}, we present such cell panels of one randomly selected cluster from MUT and WT, respectively.  With these cell panels,  we observe that cells from MUT are more circular in shape and  larger in size than those from WT group. Additionally, cells from group MUT tend to have lower intensity than those from WT. The substantial difference in cellular phenotypic features related to the shape, size, and intensity between the two groups suggests the efficacy of our complete analysis pipeline for cell phenotypic feature comparison with large-scale molecularly distinct cell populations from histopathology images.

\section{CONCLUSIONS}~\label{sec:concl}
We present a complete workflow that  facilitates quantitative histopathology imaging investigations of phenotypic feature comparison for cells from different molecular groups. The developed analysis framework consists of image color normalization, deconvolution, segmentation, feature computation, and feature comparison analysis with representative cell retrieval. As our driving use case, we test our method on histopathology microscopy images of astrocytoma brain tumors % and analyze 26,871 and 23,717 cells from IDH-mutant IDH-wildtype anaplastic astrocytomas 
from database of The Cancer Genome Atlas. With derived cellular features, we identify and retrieve representative cell instances that are phenotypically different between Isocitrate Dehydrogenase (IDH)-mutant (MUT) or wildtype (WT) groups by data pruning, dimensionality reduction, and unsupervised clustering. Our analysis is generic to a wide set of cell-based biomedical research.

% identify which point pairs to link, connect dyes, extract the region of interest
% each consecutive pair of concavity points clumped nuclei our contribution are threefold
\addtolength{\textheight}{-12cm}   % This command serves to balance the column lengths

%%%%%%%%%%%%%%%%%%%%%%%%%%%%%%%%%%%%%%%%%%%%%%%%%%%%%%%%%%%%%%%%%%%%%%%%%%%%%%%%

%%%%%%%%%%%%%%%%%%%%%%%%%%%%%%%%%%%%%%%%%%%%%%%%%%%%%%%%%%%%%%%%%%%%%%%%%%%%%%%%



%\section*{ACKNOWLEDGMENT}
%This research is supported in part by grants from NIH (K25CA181503 and R01CA176659), and CNPq.

% %%%%%%%%%%%%%%%%%%%%%%%%%%%%%%%%%%%%%%%%%%%%%%%%%%%%%%%%%%%%%%%%%%%%%%%%%%%%%%%%

% References are important to the reader; therefore, each citation must be complete and correct. If at all possible, references should be commonly available publications.

\end{document}